\documentclass[journal]{IEEEtran}
\usepackage{blindtext}
\usepackage{graphicx}
\usepackage{hyperref}
\usepackage[backend=bibtex,style=ieee,sorting=ynt]{biblatex}
\usepackage{tikz}
\usepackage{pgfplots}
\usepackage{amsmath}
\usepackage{amsfonts}
\usepackage[nolist]{acronym} 

\usepackage{amsmath,amstext,amsfonts,amssymb}
\usepackage{mathrsfs}

    \usetikzlibrary{
        pgfplots.colorbrewer,
    }
    \pgfplotsset{
        cycle list/.define={my marks}{
            every mark/.append style={solid,fill=\pgfkeysvalueof{/pgfplots/mark list fill}},mark=*\\
            every mark/.append style={solid,fill=\pgfkeysvalueof{/pgfplots/mark list fill}},mark=square*\\
            every mark/.append style={solid,fill=\pgfkeysvalueof{/pgfplots/mark list fill}},mark=triangle*\\
            every mark/.append style={solid,fill=\pgfkeysvalueof{/pgfplots/mark list fill}},mark=diamond*\\
        },
    }

\begin{document}

\title{Over the Air Deep Learning \\Based Radio Signal Classification}

\author{Tim~O'Shea,~\IEEEmembership{Senior Member,~IEEE,}
        Tamoghna~Roy,~\IEEEmembership{Member,~IEEE}\\
        and~T.~Charles~Clancy,~\IEEEmembership{Senior Member,~IEEE}%
\thanks{Authors are with the Bradley Department of Electrical and Computer Engineering, Virginia Tech and DeepSig, Arlington, VA  e-mail: (oshea,tamoghna,tcc)@vt.edu.}%
}

\maketitle

\begin{acronym}
 \acro{SIFT}{scale-invariant feature transform}
 \acro{MFCC}{Mel-frequency Cepstral coefficients}
 \acro{DL}{deep learning}
 \acro{ML}{machine learning}
 \acro{DOF}{degree of freedom}
 \acro{ACF}{auto-correlation function}
 \acro{SCF}{spectral correlation function}
 \acro{HOS}{higher order statistic}
 \acro{HOM}{higher order moment}
 \acro{HOC}{higher order cumulants}
 \acro{CFO}{carrier frequency offset}
 \acro{SRO}{symbol rate offset}
 \acro{SVM}{support vector machine}
 \acro{SGD}{stochastic gradient descent}
 \acro{RFIC}{radio frequency integrated circuit}
 \acro{USRP}{universal software radio peripheral}
 \acro{OTA}{over the air}
 \acro{LO}{local oscillator} 
 \acro{ReLU}{rectified linear unit}
 \acro{SDR}{software defined radio}
 \acro{QAM}{quadrature amplitude modulation}
 \acro{PSK}{phase shift keying}
 \acro{CNN}{convolutional neural network}
 \acro{AWGN}{additive white gaussian noise}
 \acro{SNR}{signal to noise ratio}
 \acro{FLOPS}{floating point operations per second}
 \acro{GPU}{graphics processing unit}
\end{acronym}

\begin{abstract}
We conduct an in depth study on the performance of deep learning based radio signal classification for radio communications signals.  We consider a rigorous baseline method using higher order moments and strong boosted gradient tree classification and compare performance between the two approaches across a range of configurations and channel impairments.  We consider the effects of carrier frequency offset, symbol rate, and multi-path fading in simulation and conduct over-the-air measurement of radio classification performance in the lab using software radios and compare performance and training strategies for both.  Finally we conclude with a discussion of remaining problems, and design considerations for using such techniques.
\end{abstract}

\IEEEpeerreviewmaketitle

\section{Introduction} \label{intro}

Rapidly understanding and labeling of the radio spectrum in an autonomous way is a key enabler for spectrum interference monitoring, radio fault detection, dynamic spectrum access, opportunistic mesh networking, and numerous regulatory and defense applications.  Boiling down a complex high-data rate flood of RF information to precise and accurate labels which can be acted on and conveyed compactly is a critical component today in numerous radio sensing and communications systems.  For many years, radio signal classification and modulation recognition have been accomplished by carefully hand-crafting specialized feature extractors for specific signal types and properties and by and deriving compact decision bounds from them using either analytically derived decision boundaries or statistical learned boundaries within low-dimensional feature spaces.  

In the past five years, we have seen rapid disruption occurring based on the improved neural network architectures, algorithms and optimization techniques collectively known as \ac{DL} \cite{goodfellow2016deep}. \ac{DL} has recently replaced the \ac{ML} state of the art in computer vision, voice and natural language processing; in both of these fields, feature engineering and pre-processing were once critically important topics, allowing cleverly designed feature extractors and transforms to extract pertinent information into a manageable reduced dimension representation from which labels or decisions could be readily learned with tools like support vector machines or decision trees.  Among these widely used front-end features were the \ac{SIFT} \cite{lowe2004distinctive}, the bag of words \cite{vidal2003object}, \ac{MFCC} \cite{imai1983cepstral} and others which were widely relied upon only a few years ago, but are no longer needed for state of the art performance today.

\ac{DL} greatly increased the capacity for feature learning directly on raw high dimensional input data based on high level supervised objectives due to the new found capacity for learning of very large neural network models with high numbers of free parameters.  This was made possible by the combination of strong regularization techniques \cite{srivastava2014dropout,ioffe2015batch}, greatly improved methods for \ac{SGD} \cite{tieleman2012lecture,kingma2014adam}, low cost high performance graphics card processing power, and combining of key neural network architecture innovations such as convolutional neural networks \cite{lecun1998gradient}, and rectified linear units \cite{nair2010rectified}.  It was not until Alexnet \cite{krizhevsky2012imagenet} that many of these techniques were used together to realize an increase of several orders of magnitude in the practical model size, parameter count, and target dataset and task complexity which made feature learning directly from imagery state of the art.  At this point, the trend in \ac{ML} has been relentless towards the replacement of rigid simplified analytic features and models with approximate models with much more accurate high degrees of freedom (DOF) models derived from data using end-to-end feature learning.  This trend has been demonstrated in vision, text processing, and voice, but has yet to be widely applied or fully realized on radio time series data sets until recently.

We showed in \cite{o2016convolutional, intromlcomsys} that these methods can be readily applied to simulated radio time series sample data in order to classify emitter types with excellent performance, obtaining equivalent accuracies several times more sensitive than existing best practice methods using feature based classifiers on higher order moments.  In this work we provide a more extensive dataset of additional radio signal types, a more realistic simulation of the wireless propagation environment, over the air measurement of the new dataset (i.e. real propagation effects), new methods for signal classification which drastically outperform those we initially introduced, and an in depth analysis of many practical engineering design and system parameters impacting the performance and accuracy of the radio signal classifier.

\section{Background}

\subsection{Baseline Classification Approach} \label{basemethod}

\subsubsection{Statistical Modulation Features}

For digital modulation techniques, higher order statistics and cyclo-stationary moments \cite{gardner1992signal,spooner1994robust,spooner2017modulation,abdelmutalab2016automatic,fehske2005new} are among the most widely used features to compactly sense and detect signals with strong periodic components such as are created by the structure of the carrier, symbol timing, and symbol structure for certain modulations.  By incorporating precise knowledge of this structure, expected values of peaks in \ac{ACF} and \ac{SCF} surfaces have been used successfully to provide robust classification for signals with unknown or purely random data.  For analog modulation where symbol timing does not produce these artifacts, other statistical features are useful in performing signal classification.

For our baseline features in this work, we leverage a number of compact \acp{HOS}.  To obtain these we compute the \acp{HOM} using the expression given below:

\begin{equation}
M(p,q) = E[ x^{p-q} (x^*)^q ]
\end{equation}

From these \acp{HOM} we can derive a number of \acp{HOC} which have been shown to be effective discriminators for many modulation types \cite{abdelmutalab2016automatic}.  \acp{HOC} can be computed combinatorially using \acp{HOM}, each expression varying slightly; below we show one example such expression for the $C(4,0)$ \ac{HOM}.

\begin{equation}
C(4,0) = \sqrt{M(4,0)-3\times M\left(2,0\right)^2}
\end{equation}

Additionally we consider a number of analog features which capture other statistical behaviors which can be useful, these include mean, standard deviation and kurtosis of the normalized centered amplitude, the centered phase, instantaneous frequency, absolute normalized instantaneous frequency, and several others which have shown to be useful in prior work. \cite{nandi1998algorithms}.

\subsubsection{Decision Criterion}

When mapping our baseline features to a class label, a number of compact machine learning or analytic decision processes can be used.  Probabilistically derived decision trees on expert modulation features were among the first to be used in this field, but for many years such decision processes have also been trained directly on datasets represented in their feature space.  Popular methods here include support vector machines (SVMs), decision trees (DTrees), neural networks (NNs) and ensembling methods which combine collections of classifiers to improve performance.  Among these ensembling methods are Boosting, Bagging \cite{quinlan1996bagging}, and Gradient tree boosting \cite{friedman2001greedy}.  In particular, XGBoost \cite{chen2016xgboost} has proven to be an extremely effective implementation of gradient tree boosting which has been used successfully by winners of numerous Kaggle data science competitions \cite{goldbloom2010data}.  In this work we opt to use the XGBoost approach for our feature classifier as it outperforms any single decision tree, SVM, or other method evaluated consistently as was the case in \cite{intromlcomsys}.

\subsection{Radio Channel Models}

When modeling a wireless channel there are many compact stochastic models for propagation effects which can be used \cite{goldsmith2005wireless}.  Primary impairments seen in any wireless channel include:

\begin{itemize}
    \item \ac{CFO}: carrier phase and frequency offset due to disparate \acp{LO} and motion (Doppler).
    \item \ac{SRO}: symbol clock offset and time dilation due to disparate clock sources and motion.
    \item Delay Spread: non-impulsive delay spread due to delayed reflection, diffraction and diffusion of emissions on multiple paths.
    \item Thermal Noise: additive white-noise impairment at the receiver due to physical device sensitivity.
\end{itemize}

Each of these effects can be compactly modeled well and is present in some form on any wireless propagation medium.  There are numerous additional propagation effects which can also be modeled synthetically beyond the scope of our exploration here.  

\subsection{Deep Learning Classification Approach}

\ac{DL} relies today on \ac{SGD} to optimize large parametric neural network models.  Since Alexnet \cite{krizhevsky2012imagenet} and the techniques described in section \ref{intro}, there have been numerous architectural advances within computer vision leading to significant performance improvements.  However, the core approach remains largely unchanged.  Neural networks are comprised of a series of layers which map each layer input $h_0$ to output $h_1$ using parametric dense matrix operations followed by non-linearities.  This can be expressed simply as follows, where weights, $W$, have the dimension $|h_0 \times h_1|$, bias, b, has the dimension $|h_1|$ (both constituting $\theta$), and max is applied element-wise per-output $|h_1|$ (applying \ac{ReLU} activation functions).

\begin{equation}
    h_1 = max(0, h_0 W + b)
\end{equation}

Convolutional layers can be formed by assigning a shape to inputs and outputs and forming $W$ from the replication of filter tap variables at regular strides across the input (to reduce parameter count and enforce translation invariance).

Training typically leverages a loss function ($\mathscr{L}$), in this case (for supervised classification) categorical cross-entropy, between one-hot known class labels $y_i$ (a zero vector, with a one value at the class index $i$ of the correct class) and predicted class values $\hat{y_i}$.  

\begin{equation}
\mathscr{L} (y,\hat{y}) = \frac{-1}{N} \sum_{i=0}^N \left[y_i \log(\hat{y}_i) + (1-y_i)\log(1-\hat{y}_i) \right]
\end{equation}

Back propagation of loss gradients can be used to iteratively update network weights ($\theta$) for each epoch $n$ within the network ($f(x,\theta)$) until validation loss is no longer decreasing.  We use the Adam optimizer \cite{kingma2014adam}, whose form roughly follows the conventional \ac{SGD} expression below, except for a more complex time varying expression for learning rate ($\eta$) beyond the scope of this work.

\begin{equation}  \label{eq:sgd}
\theta_{n+1} = \theta_n - \eta \frac{\partial  \mathscr{L}(y, f(x,\theta_n))}{\partial \theta_n} 
\end{equation}

To reduce over fitting to training data, regularization is used.  We use batch normalization \cite{ioffe2015batch} for regularization of convolutional layers and Alpha Dropout \cite{klambauer2017self} for regularization of fully connected layers.  Detail descriptions of additional layers used including SoftMax, Max-Pooling, etc are beyond the scope of this work and are described fully in \cite{goodfellow2016deep}.

\section{Dataset Generation Approach}

\begin{figure}[ht!]
    \centering
    \includegraphics[width=0.45\textwidth]{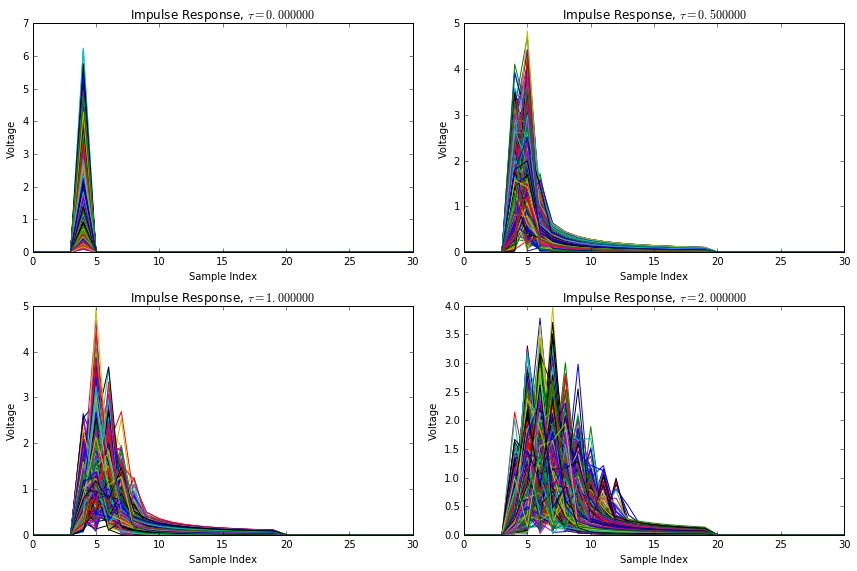}
    \caption{Fading Power Delay Profile Examples}
    \label{fig:delayspread}
\end{figure}

\begin{figure*}[ht!]
    \centering
    \includegraphics[width=1.0\textwidth]{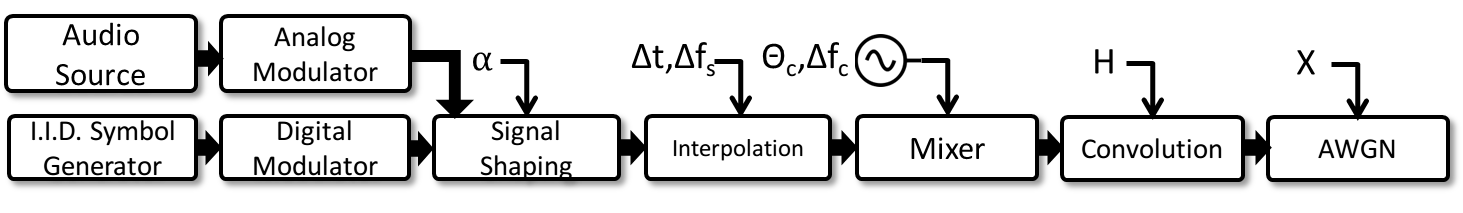}
    \caption{System for dataset signal generation and synthetic channel impairment modeling}
    \label{fig:simproc}
\end{figure*}

\begin{figure}[h]
    \centering
    \includegraphics[width=0.450\textwidth]{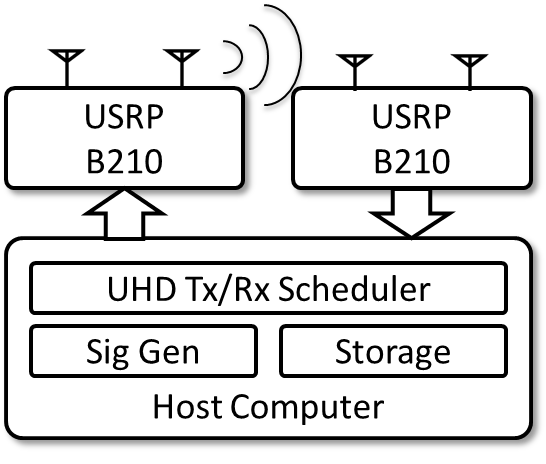}
    \caption{Over the Air Test Configuration}
    \label{fig:otadiag}
\end{figure}

\begin{figure}[h]
    \centering
    \includegraphics[width=0.450\textwidth]{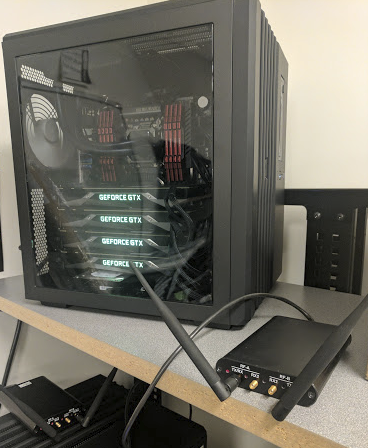}
    \caption{Configuration for Over the Air Transmission of Signals}
    \label{fig:otapic}
\end{figure}

We generate new datasets for this investigation by building upon an improved version of the tools described in \cite{o2016radio}.  24 different analog and digital modulators are used which cover a wide range of single carrier modulation schemes.  We consider several different propagation scenarios in the context of this work, first are several simulated wireless channels generated from the model shown in figure \ref{fig:simproc}, and second we consider \ac{OTA} transmission channel of clean signals as shown in figures \ref{fig:otadiag} and \ref{fig:otapic} with no synthetic channel impairments.  Digital signals are shaped with a root-raised cosine pulse shaping filter \cite{proakisdigital} with a range of roll-off values ($\alpha$).

For each example in the synthetic data sets, we independently draw a random value for each of the variables shown below in table \ref{tab:chanrnd}.  This results in a new and uncorrelated random channel initialization for each example.

\begin{table}
\renewcommand{\arraystretch}{1.0} 
\centering
\caption{Random Variable Initialization}
\label{tab:chanrnd}
\begin{tabular}{l|l}
Random Variable & Distribution \\
\hline
$\alpha $       & $ U(0.1,0.4) $ \\ 
$\Delta_t $     & $ U(0, 16) $   \\
$\Delta f_s$    & $ N(0, \sigma_{clk}) $ \\
$\theta_c $     & $ U(0, 2 \pi) $ \\
$\Delta f_c$    & $ N(0, \sigma_{clk}) $ \\
H               & $ \Sigma_i \delta(t-\text{Rayleigh}_i(\tau))  $
\end{tabular}
\end{table}

Figure \ref{fig:delayspread} illustrates several random values for $H$, the channel impulse response envelope, for different delay spreads, $\tau = [0, 0.5, 1.0, 2.0]$, relating to different levels of multi-path fading in increasingly more difficult Rayleigh fading environments.  Figure \ref{fig:const4} illustrate examples from the training set when using a simulated channel at low SNR (0 dB $E_s/N_0$).

We consider two different compositions of the dataset, first a ``Normal'' dataset, which consists of 11 classes which are all relatively low information density and are commonly seen in impaired environments.  These 11 signals represent a relatively simple classification task at high SNR in most cases, somewhat comparable to the canonical MNIST digits.  Second, we introduce a ``Difficult'' dataset, which contains all 24 modulations. These include a number of high order modulations (QAM256 and APSK256), which are used in the real world in very high-SNR low-fading channel environments such as on impulsive satellite links \cite{cioni2016transmission} (e.g. DVB-S2X).  We however, apply impairments which are beyond that which you would expect to see in such a scenario and consider only relatively short-time observation windows for classification, where the number of samples ($\ell$) is $= 1024$.  Short time classification is a hard problem since decision processes can not wait and acquire more data to increase certainty.  This is the case in many real world systems when dealing with short observations (such as when rapidly scanning a receiver) or short signal bursts in the environment.   Under these effects, with low SNR examples (from -20 dB to +30 dB $E_s/N_0$), one would not expect to be able to achieve anywhere near 100\% classification rates on the full dataset, making it a good benchmark for comparison and future research comparison.  

The specific modulations considered within each of these two dataset types are as follows: 

\begin{itemize}
    \item Normal Classes: OOK, 4ASK, BPSK, QPSK, 8PSK, 16QAM, AM-SSB-SC, AM-DSB-SC, FM, GMSK, OQPSK
    \item Difficult Classes: OOK, 4ASK, 8ASK, BPSK, QPSK, 8PSK, 16PSK, 32PSK, 16APSK, 32APSK, 64APSK, 128APSK,
                         16QAM, 32QAM, 64QAM, 128QAM, 256QAM, AM-SSB-WC, AM-SSB-SC, AM-DSB-WC, AM-DSB-SC, FM, GMSK, OQPSK
\end{itemize}

The raw datasets will be made available on the RadioML website \footnote{https://radioml.org} shortly after publication.

\subsection{Over the air data capture}

In additional to simulating wireless channel impairments, we also implement an \ac{OTA} test-bed in which we modulate and transmit signals using a \ac{USRP} \cite{ettus2015universal} B210 \ac{SDR}.  We use a second B210 (with a separate free-running \ac{LO}) to receive these transmissions in the lab, over a relatively benign indoor wireless channel on the 900MHz ISM band.  These radios use the Analog Devices AD9361 \cite{ad9361url} \ac{RFIC} as their radio front-end and have an \ac{LO} that provides a frequency (and clock) stability of around 2 parts per million (PPM).  We off-tune our signal by around 1 MHz to avoid DC signal impairment associated with direct conversion, but store signals at base-band (offset only by \ac{LO} error).  Received test emissions are stored off unmodified along with ground truth labels for the modulation from the emitter.

\section{Signal Classification Models}

In this section we explore the radio signal classification methods in more detail which we will use for the remainder of this paper.

\subsection{Baseline Method} 

Our baseline method leverages the list of higher order moments and other aggregate signal behavior statistics given in table \ref{tab:feat-map}.  Here we can compute each of these statistics over each 1024 sample example, and translate the example into feature space, a set of real values associated with each statistic for the example.  This new representation has reduced the dimension of each example from $\mathbb{R}^{1024*2}$ to $\mathbb{R}^{28}$, making the classification task much simpler but also discarding the vast majority of the data.  We use an ensemble model of gradient boosted trees (XGBoost) \cite{chen2016xgboost} to classify modulations from these features, which outperforms a single decision tree or \ac{SVM} significantly on the task.

\begin{table}
\renewcommand{\arraystretch}{1.0} 
\centering
\caption{Features Used}
\label{tab:feat-map}
\begin{tabular}{l}
 Feature Name   \\\hline
 M(2,0), M(2,1) \\
 M(4,0), M(4,1), M(4,2), M(4,3)   \\ 
 M(6,0), M(6,1), M(6,2), M(6,3)   \\ 
 C(2,0), C(2,1) \\
 C(4,0), C(4,1), C(4,2), \\
 C(6,0), C(6,1), C(6,2), C(6,3) \\ 
 Additional analog \ref{basemethod}
\end{tabular}
\end{table}

\subsection{Convolutional Neural Network}

Since \cite{lecun1998gradient} and \cite{krizhevsky2012imagenet} the use of \ac{CNN} layers to impart translation invariance in the input, followed by fully connected layers (FC) in classifiers, has been used in the computer vision problems.   In \cite{simonyan2014very}, the question of how to structure such networks is explored, and several basic design principals for "VGG" networks are introduced (e.g. filter size is minimized at 3x3, smallest size pooling operations are used at 2x2).  Following this approach has generally led to straight forward way to construct \acp{CNN} with good performance.  We adapt the VGG architecture principals to a 1D \ac{CNN}, improving upon the similar networks in \cite{o2016convolutional, intromlcomsys}.  This represents a simple \ac{DL} \ac{CNN} design approach which can be readily trained and deployed to effectively accomplish many small radio signal classification tasks.

\begin{table}
\renewcommand{\arraystretch}{1.0} 
\centering
\caption{CNN Network Layout}
\label{tab:cnn-layout}
\begin{tabular}{l|l}
 Layer    & Output dimensions    \\
 \hline
 Input & $2\times 1024$ \\
 Conv & $64 \times 1024$ \\
 Max Pool & $64 \times 512$ \\
 Conv & $64 \times 512$ \\
 Max Pool & $64 \times 256$ \\
 Conv & $64 \times 256$ \\
 Max Pool & $64 \times 128$ \\
 Conv & $64 \times 128$ \\
 Max Pool & $64 \times 64$ \\
 Conv & $64 \times 64$ \\
 Max Pool & $64 \times 32$ \\
 Conv & $64 \times 32$ \\
 Max Pool & $64 \times 16$ \\
 Conv & $64 \times 16$ \\
 Max Pool & $64 \times 8$ \\
 FC/Selu & $128$ \\
 FC/Selu & $128$ \\
 FC/Softmax & $24$ \\ 
\end{tabular}
\end{table}

Of significant note here, is that the features into this \ac{CNN} are the raw I/Q samples of each radio signal example which have been normalized to unit variance.   We do not perform any expert feature extraction or other pre-processing on the raw radio signal, instead allowing the network to learn raw time-series features directly on the high dimension data.  Real valued networks are used, as complex valued auto-differentiation is not yet mature enough for practical use.

\subsection{Residual Neural Network}

\begin{figure}[h]
    \centering
    \includegraphics[width=0.45\textwidth]{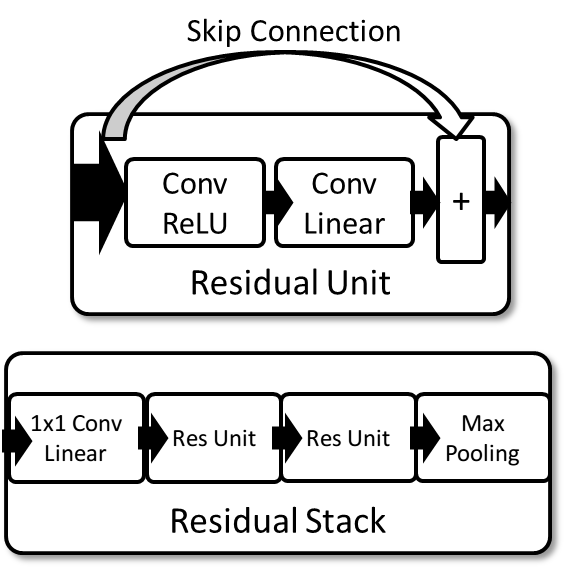}
    \caption{Hierarchical Layers Used in Network}
    \label{fig:resunit}
\end{figure}

As network algorithms and architectures have improved since Alexnet, they have made the effective training of deeper networks using more and wider layers possible, and leading to improved performance.  In our original work \cite{o2016convolutional} we employ only a small convolutional neural network with several layers to improve over the prior state of the art.  However in the computer vision space, the idea of deep residual networks has become increasingly effective \cite{he2016deep}.  In a deep residual network, as is shown in figure \ref{fig:resunit}, the notion of skip or bypass connections is used heavily, allowing for features to operate at multiple scales and depths through the network.  This has led to significant improvements in computer vision performance, and has also been used effectively on time-series audio data \cite{oord2016wavenet}.  In \cite{west2017deep}, the use of residual networks for time-series radio classification is investigated, and seen to train in fewer epochs, but not to provide significant performance improvements in terms of classification accuracy.  We revisit the problem of modulation recognition with a modified residual network and obtain improved performance when compared to the \ac{CNN} on this dataset.  The basic residual unit and stack of residual units is shown in figure \ref{fig:resunit}, while the network architecture for our best architecture for ($\ell = 1024$) is shown in table \ref{tab:resnet-layout}.  We also employ self-normalizing neural networks \cite{klambauer2017self} in the fully connected region of the network, employing the scaled exponential linear unit (SELU) activation function, mean-response scaled initialization (MRSA) \cite{he2015delving}, and Alpha Dropout \cite{klambauer2017self}, which provides a slight improvement over conventional \ac{ReLU} performance.

\begin{table}
\renewcommand{\arraystretch}{1.0} 
\centering
\caption{ResNet Network Layout}
\label{tab:resnet-layout}
\begin{tabular}{l|l}
 Layer    & Output dimensions    \\\hline
 Input & $2\times 1024$ \\
 Residual Stack & $32 \times 512$ \\
 Residual Stack & $32 \times 256$ \\
 Residual Stack & $32 \times 128$ \\
 Residual Stack & $32 \times 64$ \\
 Residual Stack & $32 \times 32$ \\
 Residual Stack & $32 \times 16$ \\
 FC/SeLU & $128$ \\
 FC/SeLU & $128$ \\
 FC/Softmax & $24$ \\
\end{tabular}
\end{table}

For the two network layouts shown, with $\ell = 1024$ and $L=5$, The ResNet has 236,344 trainable parameters, while the CNN/VGG network has a comparable 257,099 trainable parameters.

\section{Sensing Performance Analysis}

There are numerous design, deployment, training, and data considerations which can significantly effect the performance of a \ac{DL} based approach to radio signal classification which must be carefully considered when designing a solution.  In this section we explore several of the most common design parameters which impact classification accuracy including radio propagation effects, model size/depth, data set sizes, observation size, and signal modulation type. 

\subsection{Classification on Low Order Modulations}

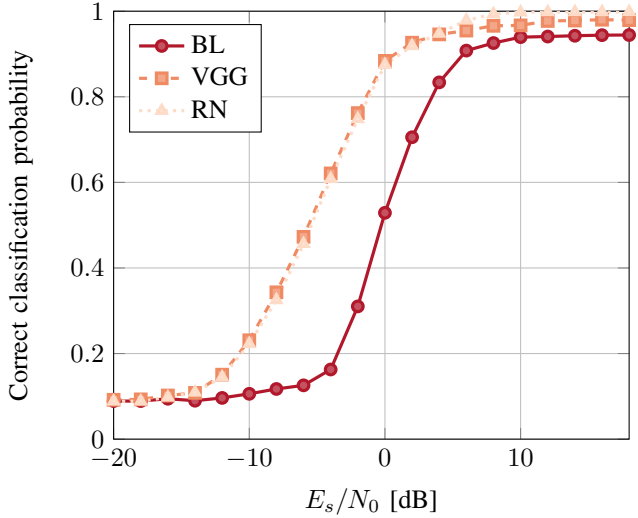
\begin{figure}
 \centering
 \begin{tikzpicture}
  \begin{axis}[
            cycle list/RdGy-6,
            mark list fill={.!75!white},
            cycle multiindex* list={
                RdGy-6 \nextlist
                my marks \nextlist
                [3 of]linestyles \nextlist
                very thick \nextlist
            },  
    xmin=-20,
    xmax=18,
    ymin=0,
    ymax=1,
    grid=major,
    xlabel={$E_s/N_0$ [dB]},
    ylabel={Correct classification probability},
    legend pos=north west,
    legend cell align={left},
    ]   
    \addplot table [x=snr, y=expert, col sep=comma]{easycase1M.csv};
    \addplot table [x=snr, y=vgg, col sep=comma]{easycase1M.csv};
    \addplot table [x=snr, y=resnet, col sep=comma]{easycase1M.csv};
    \legend{{BL},{VGG},{RN}}
    
  \end{axis}
 \end{tikzpicture}
 \caption{11-modulation AWGN dataset performance comparison (N=1M)}\label{fig:awgnperf}
\end{figure}

We first compare performance on the lower difficulty dataset on lower order modulation types.  Training on a dataset of 1 million example, each 1024 samples long, we obtain excellent performance at high SNR for both the VGG \ac{CNN} and the ResNet (RN) \ac{CNN}.

In this case, the ResNet achieves roughly 5 dB higher sensitivity for equivalent classification accuracy than the baseline, and at high SNR a maximum classification accuracy rate of 99.8\% is achieved by the ResNet, while the VGG network achieves 98.3\% and the baseline method achieves a 94.6\% accuracy.  At lower SNRs, performance between VGG and ResNet networks are virtually identical, but at high-SNR performance improves considerably using the ResNet and obtaining almost perfect classification accuracy.

For the remainder of the paper, we will consider the much harder task of 24 class high order modulations containing higher information rates and much more easily confused classes between multiple high order PSKs, APSKs and QAMs.

\subsection{Classification under AWGN conditions}

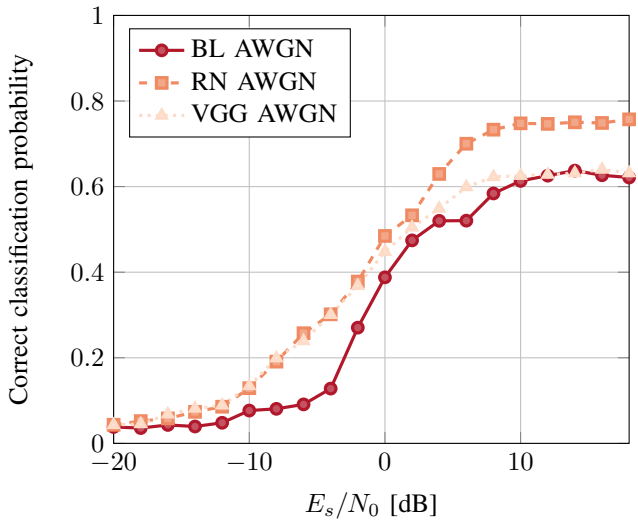
\begin{figure}
 \centering
 \begin{tikzpicture}
  \begin{axis}[
            cycle list/RdGy-6,
            mark list fill={.!75!white},
            cycle multiindex* list={
                RdGy-6 \nextlist
                my marks \nextlist
                [3 of]linestyles \nextlist
                very thick \nextlist
            },  
    xmin=-20,
    xmax=18,
    ymin=0,
    ymax=1,
    grid=major,
    xlabel={$E_s/N_0$ [dB]},
    ylabel={Correct classification probability},
    legend pos=north west,
    legend cell align={left},
    ]   
    \addplot table [x=snr, y=baseline_AWGN, col sep=comma]{baselineimp.csv};
    \addplot table [x=snr, y=resnet_AWGN_rf, col sep=comma]{impairment.csv};
    \addplot table [x=snr, y=vgg_AWGN, col sep=comma]{vggimp.csv};
    \legend{{BL AWGN},{RN AWGN},{VGG AWGN},
    }
  \end{axis}
 \end{tikzpicture}
 \caption{Comparison models under AWGN (N=240k)}\label{fig:awgnperf}
\end{figure}

Signal classification under \ac{AWGN} is the canonical problem which has been explored for many years in communications literature.  It is a simple starting point, and it is the condition under which analytic feature extractors should generally perform their best (since they were derived under these conditions).  In figure \ref{fig:awgnperf} we compare the performance of the ResNet (RN), VGG network, and the baseline (BL) method on our full dataset for $\ell=1024$ samples, $N=239,616$ examples, and $L=6$ residual stacks.  Here, the residual network provides the best performance at both high and low SNRs on the difficult dataset by a margin of 2-6 dB in improved sensitivity for equivalent classification accuracy.  

\subsection{Classification under Impairments}

In any real world scenario, wireless signals are impaired by a number of effects.  While AWGN is widely used in simulation and modeling, the effects described above are present almost universally.  It is interesting to inspect how well learned classifiers perform under such impairments and compare their rate of degradation under these impairments with that of more traditional approaches to signal classification.

\begin{figure}
 \centering
 \begin{tikzpicture}
  \begin{axis}[
            cycle list/RdGy-6,
            mark list fill={.!75!white},
            cycle multiindex* list={
                RdGy-6 \nextlist
                my marks \nextlist
                [3 of]linestyles \nextlist
                very thick \nextlist
            },  
    xmin=-20,
    xmax=18,
    ymin=0,
    ymax=1,
    grid=major,
    xlabel={$E_s/N_0$ [dB]},
    ylabel={Correct classification probability},
    legend pos=north west,
    legend cell align={left},
    ]   
    \addplot table [x=snr, y=resnet_AWGN_rf, col sep=comma]{impairment.csv};
    \addplot table [x=snr, y=resnet_OSC.0001_rf, col sep=comma]{impairment.csv};
    \addplot table [x=snr, y=resnet_OSC.01_rf, col sep=comma]{impairment.csv};
    \addplot table [x=snr, y=resnet_FADE.5_rf, col sep=comma]{impairment.csv};
    \addplot table [x=snr, y=resnet_FADE1_rf, col sep=comma]{impairment.csv};
    \addplot table [x=snr, y=resnet_FADE2_rf, col sep=comma]{impairment.csv};
    \addplot table [x=snr, y=resnet_FADE4_rf, col sep=comma]{impairment.csv};
    \legend{{RN AWGN},{RN $\sigma_{clk}=0.01$},{RN $\sigma_{clk}=0.0001$},
            {RN $\tau=0.5$},{RN $\tau=1$},{RN $\tau=2$},{RN $\tau=4$},
    }
  \end{axis}
 \end{tikzpicture}
 \caption{Resnet performance under various channel impairments (N=240k)}\label{fig:perfimpresnet}
\end{figure}
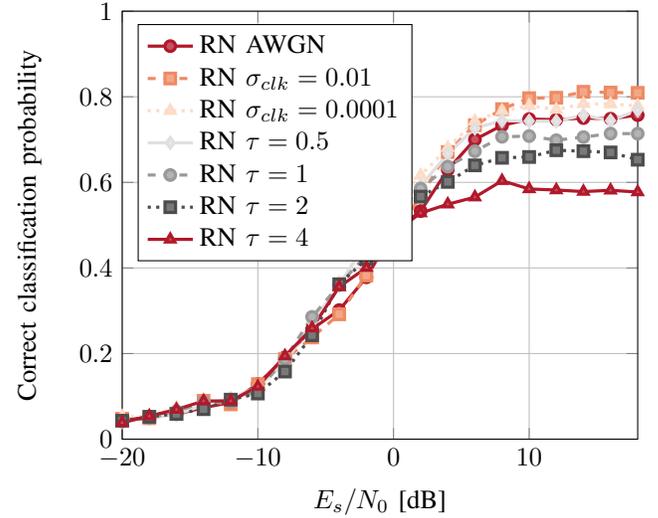

In figure \ref{fig:perfimpresnet} we plot the performance of the residual network based classifier under each considered impairment model.  This includes AWGN, $\sigma_{clk}=0.0001$ - minor LO offset, $\sigma_{clk}=0.01$ - moderate LO offset, and several fading models ranging from $\tau=0.5$ to $\tau=4.0$.  Under the fading models, moderate LO offset is assumed as well.  Interestingly in this plot, ResNet performance improves under LO offset rather than degrading.  Additional LO offset which results in spinning or dilated versions of the original signal, appears to have a positive regularizing effect on the learning process which provides quite a noticeable improvement in performance.  At high SNR performance ranges from around 80\% in the best case down to about 59\% in the worst case.

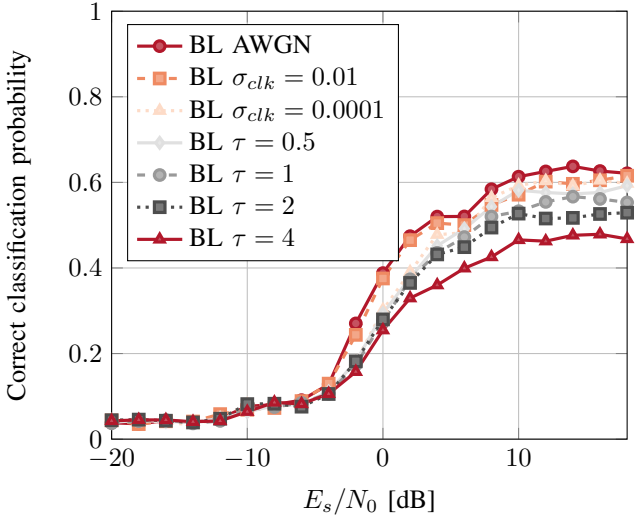
\begin{figure}
 \centering
 \begin{tikzpicture}
  \begin{axis}[
            cycle list/RdGy-6,
            mark list fill={.!75!white},
            cycle multiindex* list={
                RdGy-6 \nextlist
                my marks \nextlist
                [3 of]linestyles \nextlist
                very thick \nextlist
            },  
    xmin=-20,
    xmax=18,
    ymin=0,
    ymax=1,
    grid=major,
    xlabel={$E_s/N_0$ [dB]},
    ylabel={Correct classification probability},
    legend pos=north west,
    legend cell align={left},
    ]   
    \addplot table [x=snr, y=baseline_AWGN, col sep=comma]{baselineimp.csv};
    \addplot table [x=snr, y=baseline_OSC.0001, col sep=comma]{baselineimp.csv};
    \addplot table [x=snr, y=baseline_OSC.01, col sep=comma]{baselineimp.csv};
    \addplot table [x=snr, y=baseline_FADE.5, col sep=comma]{baselineimp.csv};
    \addplot table [x=snr, y=baseline_FADE1, col sep=comma]{baselineimp.csv};
    \addplot table [x=snr, y=baseline_FADE2, col sep=comma]{baselineimp.csv};
    \addplot table [x=snr, y=baseline_FADE4, col sep=comma]{baselineimp.csv};
    \legend{{BL AWGN},{BL $\sigma_{clk}=0.01$},{BL $\sigma_{clk}=0.0001$},
            {BL $\tau=0.5$},{BL $\tau=1$},{BL $\tau=2$},{BL $\tau=4$},
    }
  \end{axis}
 \end{tikzpicture}
 \caption{Baseline performance under channel impairments (N=240k) }\label{fig:perfimpbase}
\end{figure}

In figure \ref{fig:perfimpbase} we show the degradation of the baseline classifier under impairments.  In this case, LO offset never helps, but the performance instead degrades with both LO offset and fading effects, in the best case at high SNR this method obtains about 61\% accuracy while in the worst case it degrades to around 45\% accuracy. 

\begin{figure}
 \centering
 \begin{tikzpicture}
  \begin{axis}[
            cycle list/RdGy-6,
            mark list fill={.!75!white},
            cycle multiindex* list={
                RdGy-6 \nextlist
                my marks \nextlist
                [3 of]linestyles \nextlist
                very thick \nextlist
            },  
    xmin=-20,
    xmax=18,
    ymin=0,
    ymax=1,
    grid=major,
    xlabel={$E_s/N_0$ [dB]},
    ylabel={Correct classification probability},
    legend pos=north west,
    legend cell align={left},
    ]   
    \addplot table [x=snr, y=baseline_OSC.01, col sep=comma]{baselineimp.csv};
    \addplot table [x=snr, y=resnet_OSC.01_rf, col sep=comma]{impairment.csv};
    \addplot table [x=snr, y=vgg_OSC.01, col sep=comma]{vggimp.csv};
    \legend{{BL $\sigma_{clk}=0.01$},{RN $\sigma_{clk}=0.01$},{VGG $\sigma_{clk}=0.01$}
    }
  \end{axis}
 \end{tikzpicture}
 \caption{Comparison models under LO impairment}\label{fig:perfimpbest}
\end{figure}
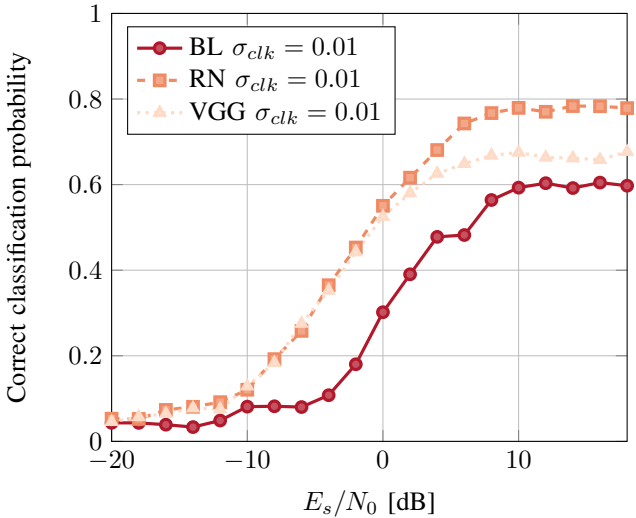

Directly comparing the performance of each model under moderate LO impairment effects, in figure \ref{fig:perfimpbest} we show that for many real world systems with unsynchronized \acp{LO} and Doppler frequency offset there is nearly a 6dB performance advantage of the ResNet approach vs the baseline, and a 20\% accuracy increase at high SNR.   In this section, all models are trained using $N=239,616$ and $\ell=1024$ for this comparison.

\subsection{Classifier performance by depth}

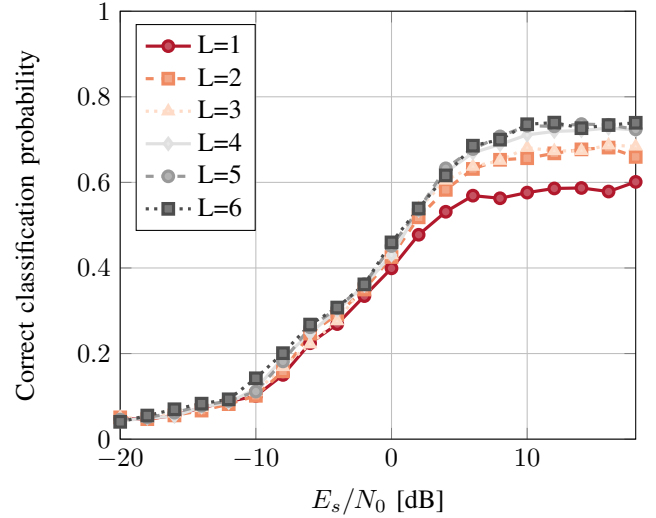
\begin{figure}
 \centering
 \begin{tikzpicture}
  \begin{axis}[
            cycle list/RdGy-6,
            mark list fill={.!75!white},
            cycle multiindex* list={
                RdGy-6 \nextlist
                my marks \nextlist
                [3 of]linestyles \nextlist
                very thick \nextlist
            },  
    xmin=-20,
    xmax=18,
    ymin=0,
    ymax=1,
    grid=major,
    xlabel={$E_s/N_0$ [dB]},
    ylabel={Correct classification probability},
    legend pos=north west,
    legend cell align={left},
    ]   
    \addplot table [x=snr, y=l1, col sep=comma]{depth.csv};
    \addplot table [x=snr, y=l2, col sep=comma]{depth.csv};
    \addplot table [x=snr, y=l3, col sep=comma]{depth.csv};
    \addplot table [x=snr, y=l4, col sep=comma]{depth.csv};
    \addplot table [x=snr, y=l5, col sep=comma]{depth.csv};
    \addplot table [x=snr, y=l6, col sep=comma]{depth.csv};
    \legend{{L=1},{L=2},{L=3},{L=4},{L=5},{L=6}}
  \end{axis}
 \end{tikzpicture}
 \caption{ResNet performance vs depth (L = number of residual stacks)}\label{fig:perfdepth}
\end{figure}

Model size can have a significant impact on the ability of large neural network models to accurately represent complex features.  In computer vision, convolutional layer based \ac{DL} models for the ImageNet dataset started around 10 layers deep, but modern state of the art networks on ImageNet are often over 100 layers deep \cite{szegedy2015going}, and more recently even over 200 layers.   Initial investigations of deeper networks in \cite{west2017deep} did not show significant gains from such large architectures, but with use of deep residual networks on this larger dataset, we begin to see quite a benefit to additional depth.  This is likely due to the significantly larger number of examples and classes used.  In figure \ref{fig:perfdepth} we show the increasing validation accuracy of deep residual networks as we introduce more residual stack units within the network architecture (i.e. making the network deeper).  We see that performance steadily increases with depth in this case with diminishing returns as we approach around 6 layers.  When considering all of the primitive layers within this network, when $L=6$ we the ResNet has 121 layers and 229k trainable parameters, when $L=0$ it has 25 layers and 2.1M trainable parameters.   Results are shown for $N=239,616$ and $\ell=1024$.

\subsection{Classification performance by modulation type}

\begin{figure*}
 \centering
 \begin{tikzpicture}
 \pgfplotsset{width=14cm}
  \begin{axis}[
              cycle list/RdGy-6,
            mark list fill={.!75!white},
            cycle multiindex* list={
                RdGy-6 \nextlist
                very thick \nextlist                
            },  
    legend style={at={(1.26,0.98)}},
    xmin=-20,
    xmax=18,
    ymin=0,
    ymax=1,
    grid=major,
    xlabel={Signal to noise ratio ($E_s/N_0$) [dB]},
    ylabel={Correct classification probability},
    legend cell align={left},
    ]   
    \addplot [color=black, mark=*,very thick] table [x=snr, y=OOK, col sep=comma]{modperf2.csv};
    \addplot [color=blue, mark=* ,very thick] table [x=snr, y=4ASK, col sep=comma]{modperf2.csv};
    \addplot [color=red, mark=*  ,very thick] table [x=snr, y=8ASK, col sep=comma]{modperf2.csv};
    \addplot [color=green, mark=*,very thick] table [x=snr, y=BPSK, col sep=comma]{modperf2.csv};
    \addplot [color=orange, mark=*,very thick] table [x=snr, y=QPSK, col sep=comma]{modperf2.csv};
    \addplot [color=black!40, mark=*,very thick] table [x=snr, y=8PSK, col sep=comma]{modperf2.csv};
    
    \addplot [color=black, mark=square,very thick] table [x=snr, y=16PSK, col sep=comma]{modperf2.csv};
    \addplot [color=blue, mark=square,very thick] table [x=snr, y=32PSK, col sep=comma]{modperf2.csv};
    \addplot [color=red, mark=square,very thick] table [x=snr, y=16APSK, col sep=comma]{modperf2.csv};
    \addplot [color=green, mark=square,very thick] table [x=snr, y=32APSK, col sep=comma]{modperf2.csv};
    \addplot [color=orange, mark=square,very thick] table [x=snr, y=64APSK, col sep=comma]{modperf2.csv};
    \addplot [color=black!40, mark=square,very thick] table [x=snr, y=128APSK, col sep=comma]{modperf2.csv};
    
    \addplot [color=black, dashed, mark=triangle,very thick] table [x=snr, y=16QAM, col sep=comma]{modperf2.csv};
    \addplot [color=blue, dashed, mark=triangle,very thick] table [x=snr, y=32QAM, col sep=comma]{modperf2.csv};
    \addplot [color=red, dashed, mark=triangle,very thick] table [x=snr, y=64QAM, col sep=comma]{modperf2.csv};
    \addplot [color=green, dashed, mark=triangle,very thick] table [x=snr, y=128QAM, col sep=comma]{modperf2.csv};
    \addplot [color=orange, dashed, mark=triangle,very thick] table [x=snr, y=256QAM, col sep=comma]{modperf2.csv};
    \addplot [color=black!40, dashed, mark=triangle,very thick] table [x=snr, y=AM-SSB-WC, col sep=comma]{modperf2.csv};
    
    \addplot [color=black,very thick] table [x=snr, y=AM-SSB-SC, col sep=comma]{modperf2.csv};
    \addplot [color=blue,very thick] table [x=snr, y=AM-DSB-WC, col sep=comma]{modperf2.csv};
    \addplot [color=red,very thick] table [x=snr, y=AM-DSB-SC, col sep=comma]{modperf2.csv};
    \addplot [color=green,very thick] table [x=snr, y=FM, col sep=comma]{modperf2.csv};
    \addplot [color=orange,very thick] table [x=snr, y=GMSK, col sep=comma]{modperf2.csv};
    \addplot [color=black!40,very thick] table [x=snr, y=OQPSK, col sep=comma]{modperf2.csv};
    
    \legend{
        {OOK},{4ASK},{8ASK},{BPSK},
        {QPSK},{8PSK},{16PSK},{32PSK},
        {16APSK},{32APSK},{64APSK},{128APSK},
        {16QAM},{32QAM},{64QAM},{128QAM},
        {256QAM},{AM-SSB-WC},{AM-SSB-SC},{AM-DSB-WC},
        {AM-DSB-SC},{FM},{GMSK},{OQPSK}
    }
  \end{axis}
 \end{tikzpicture}
 \caption{Modrec performance vs modulation type (Resnet on synthetic data with N=1M, $\sigma_{clk}$=0.0001)}\label{fig:modperf}
\end{figure*}
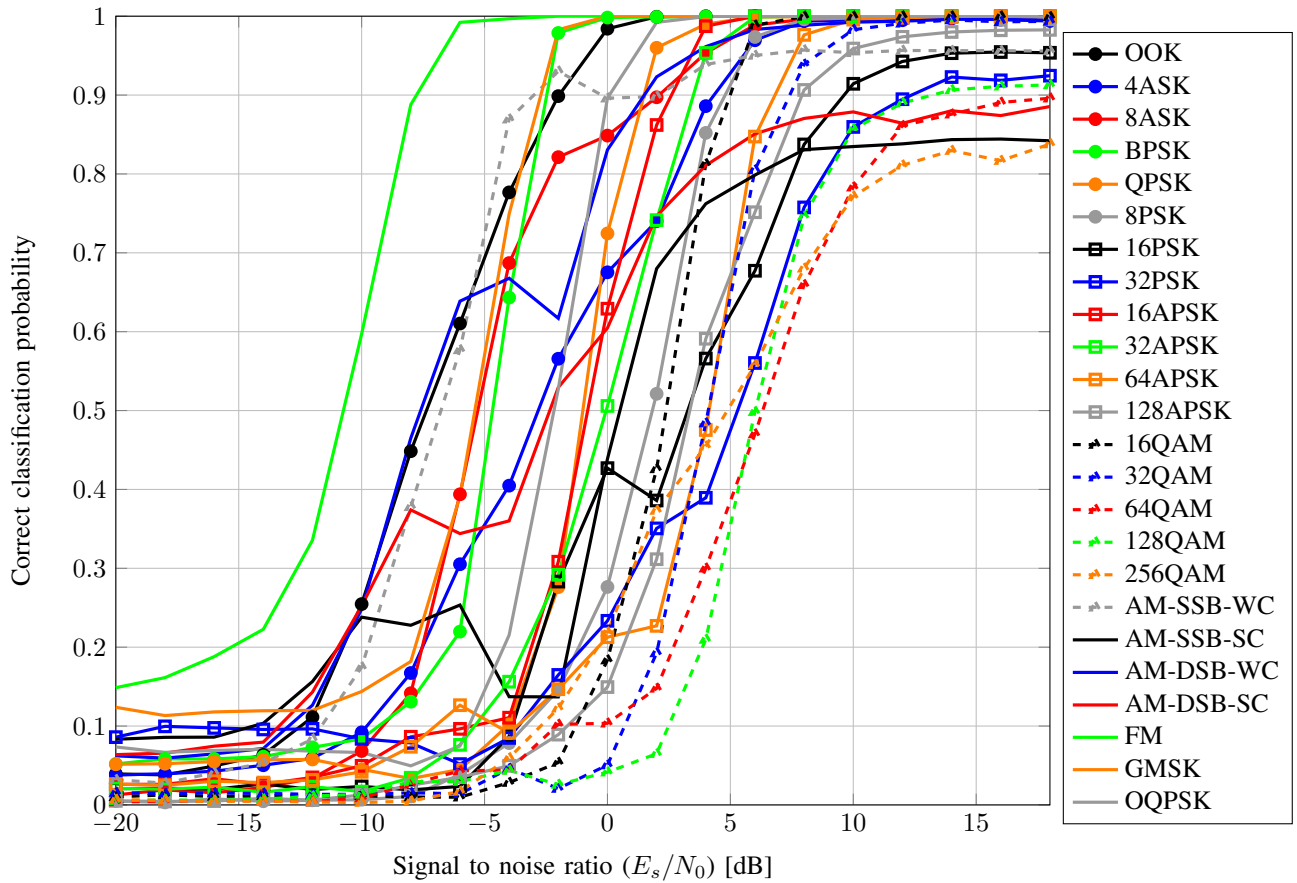

\begin{figure}[h]
    \centering
    \includegraphics[width=0.45\textwidth]{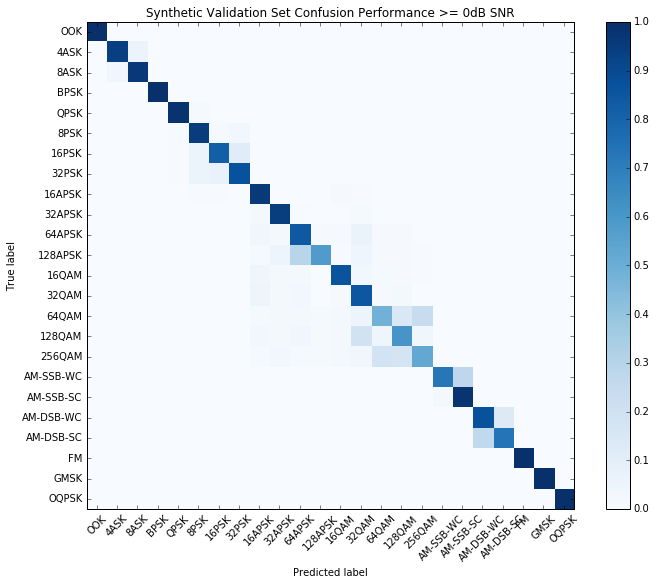}
    \caption{24-modulation confusion matrix for ResNet trained and tested on synthetic dataset with N=1M and \ac{AWGN}}
    \label{fig:confawgn24}
\end{figure}

In figure \ref{fig:modperf} we show the performance of the classifier for individual modulation types.  Detection performance of each modulation type varies drastically over about 18dB of \ac{SNR}.  Some signals with lower information rates and vastly different structure such as AM and FM analog modulations are much more readily identified at low SNR, while high-order modulations require higher \acp{SNR} for robust performance and never reach perfect classification rates.  However, all modulation types reach rates above 80\% accuracy by around 10dB \ac{SNR}.   In figure \ref{fig:confawgn24} we show a confusion matrix for the classifier across all 24 classes for AWGN validation examples where \ac{SNR} is greater than or equal to zero.  We can see again here that the largest sources of error are between high order \ac{PSK} (16/32-PSK), between high order \ac{QAM} (64/128/256-QAM), as well as between AM modes (confusing with-carrier (WC) and suppressed-carrier (SC)).  This is largely to be expected as for short time observations, and under noisy observations, high order \ac{QAM} and \ac{PSK} can be extremely difficult to tell apart through any approach.

\subsection{Classifier Training Size Requirements}

\begin{figure}
 \centering
 \begin{tikzpicture}
  \begin{axis}[
          legend style={font=\small},
            cycle list/RdGy-6,
            mark list fill={.!75!white},
            cycle multiindex* list={
                RdGy-6 \nextlist
                my marks \nextlist
                [3 of]linestyles \nextlist
                very thick \nextlist
            },  
    xmin=-20,
    xmax=18,
    ymin=0,
    ymax=1,
    grid=major,
    xlabel={$E_s/N_0$ [dB]},
    ylabel={Correct classification probability},
    legend pos=north west,
    legend cell align={left},
    ]   
    \addplot table [x=snr, y=s976, col sep=comma]{nexamp.csv};
    \addplot table [x=snr, y=s1953, col sep=comma]{nexamp.csv};
    \addplot table [x=snr, y=s3906, col sep=comma]{nexamp.csv};
    \addplot table [x=snr, y=s7812, col sep=comma]{nexamp.csv};
    \addplot table [x=snr, y=s15625, col sep=comma]{nexamp.csv};
    \addplot table [x=snr, y=s31250, col sep=comma]{nexamp.csv};
    \addplot table [x=snr, y=s62500, col sep=comma]{nexamp.csv};
    \addplot table [x=snr, y=s125000, col sep=comma]{nexamp.csv};
    \addplot table [x=snr, y=s250000, col sep=comma]{nexamp.csv};
    \addplot table [x=snr, y=s500000, col sep=comma]{nexamp.csv};
    \addplot table [x=snr, y=s1000000, col sep=comma]{nexamp.csv};
    \addplot table [x=snr, y=s2000000, col sep=comma]{nexamp.csv};
    \legend{{N=1k},{N=2k},{N=4k},{N=8k},{N=15k},{N=31k},{N=62k},{N=125k},{N=250k},{N=500k},{N=1M},{N=2M}}
  \end{axis}
 \end{tikzpicture}
 \caption{Performance vs training set size (N) with $\ell$ = 1024}\label{fig:perfexamples}
\end{figure}

\begin{figure}[h]
    \centering
    \includegraphics[width=0.45\textwidth]{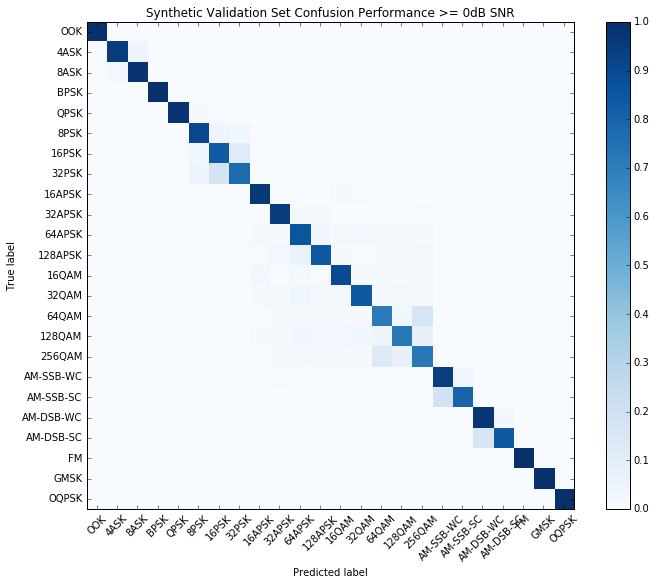}
    \caption{24-modulation confusion matrix for ResNet trained and tested on synthetic dataset with N=1M and $\sigma_{clk}=0.0001$}
    \label{fig:confsim}
\end{figure}

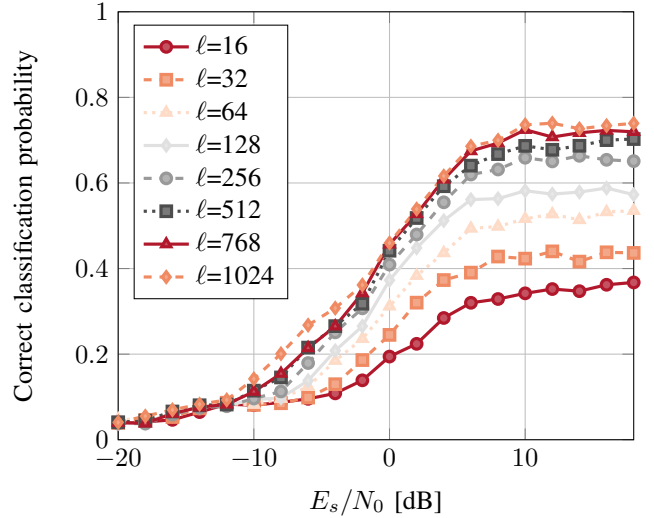
\begin{figure}
 \centering
 \begin{tikzpicture}
  \begin{axis}[
            cycle list/RdGy-6,
            mark list fill={.!75!white},
            cycle multiindex* list={
                RdGy-6 \nextlist
                my marks \nextlist
                [3 of]linestyles \nextlist
                very thick \nextlist
            },  
    xmin=-20,
    xmax=18,
    ymin=0,
    ymax=1,
    grid=major,
    xlabel={$E_s/N_0$ [dB]},
    ylabel={Correct classification probability},
    legend pos=north west,
    legend cell align={left},
    ]   
    \addplot table [x=snr, y=s16, col sep=comma]{nsamp.csv};
    \addplot table [x=snr, y=s32, col sep=comma]{nsamp.csv};
    \addplot table [x=snr, y=s64, col sep=comma]{nsamp.csv};
    \addplot table [x=snr, y=s128, col sep=comma]{nsamp.csv};
    \addplot table [x=snr, y=s256, col sep=comma]{nsamp.csv};
    \addplot table [x=snr, y=s512, col sep=comma]{nsamp.csv};
    \addplot table [x=snr, y=s768, col sep=comma]{nsamp.csv};
    \addplot table [x=snr, y=s1024, col sep=comma]{nsamp.csv};
    \legend{{$\ell$=16},{$\ell$=32},{$\ell$=64},{$\ell$=128},
            {$\ell$=256},{$\ell$=512},{$\ell$=768},{$\ell$=1024}}
  \end{axis}
 \end{tikzpicture}
 \caption{Performance vs example length in samples ($\ell$)}\label{fig:perfsamples}
\end{figure}

When using data-centric machine learning methods, the dataset often has an enormous impact on the quality of the model learned.  We consider the influence of the number of example signals in the training set, $N$, as well as the time-length of each individual example in number of samples, $\ell$.  

In figure \ref{fig:perfexamples} we show how performance of the resulting model changes based on the total number of training examples used.  Here we see that dataset size has a dramatic impact on model training, high \ac{SNR} classification accuracy is near random until 4-8k examples and improves 5-20\% with each doubling until around 1M.  These results illustrate that having sufficient training data is critical for performance.  For the largest case, with 2 million examples, training on a single state of the art Nvidia V100 \ac{GPU} (with approximately 125 tera-\ac{FLOPS}) takes around 16 hours to reach a stopping point, making significant experimentation at these dataset sizes cumbersome.  We do not see significant improvement going from 1M to 2M examples, indicating a point of diminishing returns for number of examples around 1M with this configuration.  With either 1M or 2M examples we obtain roughly 95\% test set accuracy at high \ac{SNR}.  The class-confusion matrix for the best performing mode with $\ell$=1024 and N=1M is shown in figure \ref{fig:confsim} for test examples at or above 0dB SNR, in all instances here we use the $\sigma_{clk}=0.0001$ dataset, which yields slightly better performance than AWGN.

Figure \ref{fig:perfsamples} shows how the model performance varies by window size, or the number of time-samples per example used for a single classification.  Here we obtain approximately a 3\% accuracy improvement for each doubling of the input size (with N=240k), with significant diminishing returns once we reach $\ell=512$ or $\ell=1024$.  We find that \acp{CNN} scale very well up to this 512-1024 size, but may need additional scaling strategies thereafter for larger input windows simply due to memory requirements, training time requirements, and dataset requirements.

\subsection{Over the air performance}

\begin{figure}[ht!]
    \centering
    \includegraphics[width=0.45\textwidth]{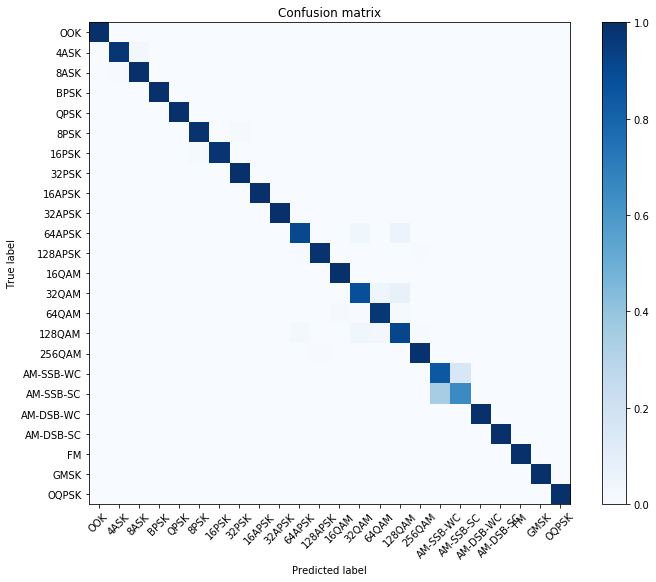}
    \caption{24-modulation confusion matrix for ResNet trained and tested on OTA examples with SNR $\sim$ 10 dB}
    \label{fig:otaconf}
\end{figure}

We generate 1.44M examples of the 24 modulation dataset over the air using the USRP setup described above.  Using a partition of 80\% training and 20\% test, we can directly train a ResNet for classification.  Doing so on an Nvidia V100 in around 14 hours, we obtain a 95.6\% test set accuracy on the over the air dataset, where all examples are roughly 10dB SNR.  A confusion matrix for this \ac{OTA} test set performance based on direct training is shown in figure \ref{fig:otaconf}.

\subsection{Transfer learning over-the-air performance}

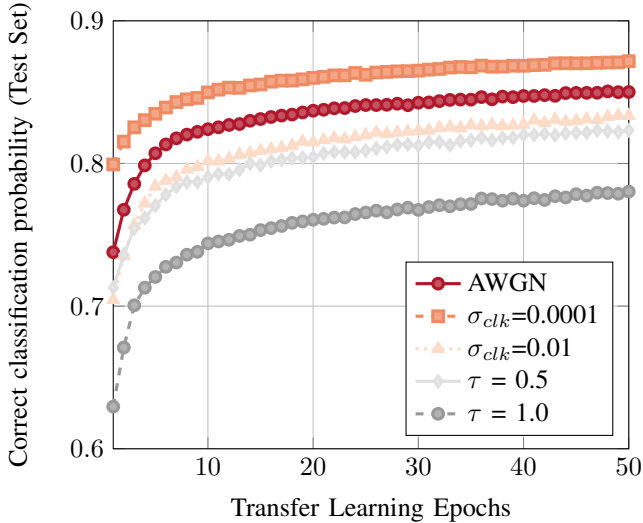
\begin{figure}
 \centering
 \begin{tikzpicture}
  \begin{axis}[
            cycle list/RdGy-6,
            mark list fill={.!75!white},
            cycle multiindex* list={
                RdGy-6 \nextlist
                my marks \nextlist
                [3 of]linestyles \nextlist
                very thick \nextlist
            },  
    xmin=1,
    xmax=50,
    ymin=.6,
    ymax=.9,
    grid=major,
    xlabel={Transfer Learning Epochs},
    ylabel={Correct classification probability (Test Set)},
    legend pos=south east,
    legend cell align={left},
    ]   
    \addplot table [x=epoch, y=AWGN, col sep=comma]{transfer.csv};
    \addplot table [x=epoch, y=OSC.0001, col sep=comma]{transfer.csv};
    \addplot table [x=epoch, y=OSC.01, col sep=comma]{transfer.csv};
    \addplot table [x=epoch, y=FADE.5, col sep=comma]{transfer.csv};
    \addplot table [x=epoch, y=FADE1, col sep=comma]{transfer.csv};
    \legend{{AWGN},{$\sigma_{clk}$=0.0001},{$\sigma_{clk}$=0.01},{$\tau$ = 0.5},{$\tau$ = 1.0}}
  \end{axis}
 \end{tikzpicture}
 \caption{RESNET Transfer Learning OTA Performance (N=120k)} \label{fig:transfer}
\end{figure}

We also consider over the air signal classification as a transfer learning problem, where the model is trained on synthetic data and then only evaluated and/or fine-tuned on \ac{OTA} data.  Because full model training can take hours on a high end \ac{GPU} and typically requires a large dataset to be effective, transfer learning is a convenient alternative for leveraging existing models and updating them on smaller computational platforms and target datasets.  We consider transfer learning, where we freeze network parameter weights for all layers except the last several fully connected layers (last three layers from table \ref{tab:resnet-layout}) in our network when while updating.  This is commonly done today with computer vision models where it is common start by using pre-trained VGG or other model weights for ImageNet or similar datasets and perform transfer learning using another dataset or set of classes.  In this case, many low-level features work well for different classes or datasets, and do not need to change during fine tuning.  In our case, we consider several cases where we start with models trained on simulated wireless impairment models using residual networks and then evaluate them on \ac{OTA} examples.  The accuracies of our initial models (trained with N=1M) on synthetic data shown in figure \ref{fig:perfimpresnet}, and these ranged from 84\% to 96\% on the hard 24-class dataset.  Evaluating performance of these models on \ac{OTA} data, without any model updates, we obtain classification accuracies between 64\% and 80\%.  By fine-tuning the last two layers of these models on the \ac{OTA} data using transfer learning, we and can recover approximately 10\% of additional accuracy.  The validation accuracies are shown for this process in figure \ref{fig:transfer}.  These ResNet update epochs on dense layers for 120k examples take roughly 60 seconds on a Titan X card to execute instead of the full $\sim$ 500 seconds on V100 card per epoch when updating model weights. 

\begin{figure}[ht!]
    \centering
    \includegraphics[width=0.45\textwidth]{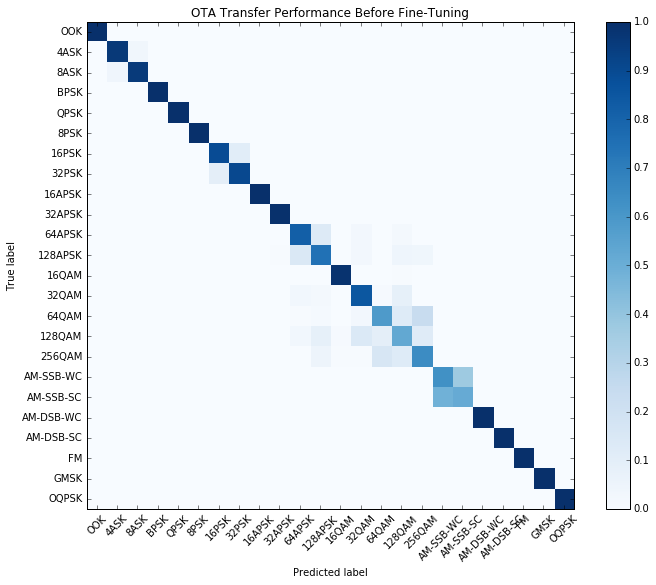}
    \caption{24-modulation confusion matrix for ResNet trained on synthetic $\sigma_{clk}=0.0001$ and tested on OTA examples with SNR $\sim$ 10 dB (prior to fine-tuning)}
    \label{fig:otatransconf}
\end{figure}
\begin{figure}[ht!]
    \centering
    \includegraphics[width=0.45\textwidth]{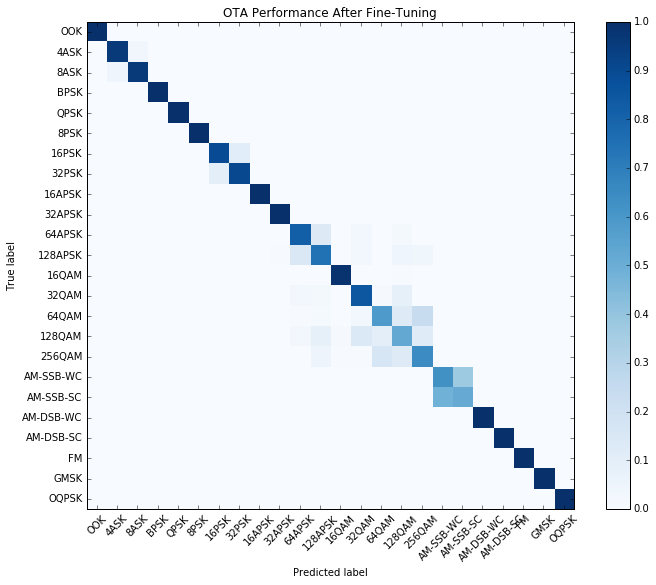}
    \caption{24-modulation confusion matrix for ResNet trained on synthetic $\sigma_{clk}=0.0001$ and tested on OTA examples with SNR $\sim$ 10 dB (after fine-tuning)}
    \label{fig:otatransconfafter}
\end{figure}

Ultimately, the model trained on just moderate \ac{LO} offset ($\sigma_{clk}$ = 0.0001) performs the best on \ac{OTA} data.  The model obtained 94\% accuracy on synthetic data, and drops roughly 7\% accuracy when evaluating on \ac{OTA} data, obtaining an accuracy of 87\%.  The primary confusion cases prior to training seem to be dealing with suppress or non-suppressed carrier analog signals, as well as the high order QAM and APSK modes.

This seems like it is perhaps the best suited among our models to match the \ac{OTA} data.  Very small \ac{LO} impairments are present in the data, the radios used had extremely stable oscillators present (GPSDO modules providing high stable \~75 PPB clocks) over very short example lengths (1024 samples), and that the two radios were essentially right next to each other, providing a very clean impulsive direct path while any reflections from the surrounding room were likely significantly attenuated in comparison, making for a near impulsive channel.  Training on harsher impairments seemed to degrade performance of the \ac{OTA} data significantly.  

We suspect as we evaluate the performance of the model under increasingly harsh real world scenarios, our transfer learning will favor synthetic models which are similarly impaired and most closely match the real wireless conditions (e.g. matching LO distributions, matching fading distributions, etc).  In this way, it will be important for this class of systems to train either directly on target signal environments, or on very good impairment simulations of them under which well suited models can be derived.  Possible mitigation to this are to include domain-matched attention mechanisms such as the radio transformer network \cite{o2016radio} in the network architecture to improve generalization to varying wireless propagation conditions.

\section{Discussion}

In this work we have extended prior work on using deep convolutional neural networks for radio signal classification by heavily tuning deep residual networks for the same task.  We have also conducted a much more thorough set of performance evaluations on how this type of classifier performs over a wide range of design parameters, channel impairment conditions, and training dataset parameters.  This residual network approach achieves state of the art modulation classification performance on a difficult new signal database both synthetically and in over the air performance.  Other architectures still hold significant potential, radio transformer networks, recurrent units, and other approaches all still need to be adapted to the domain, tuned and quantitatively benchmarked against the same dataset in the future.  Other works have explored these to some degree, but generally not with sufficient hyper-parameter optimization to be meaningful.  

We have shown that, contrary to prior work, deep networks do provide significant performance gains for time-series radio signals where the need for such deep feature hierarchies was not apparent, and that residual networks are a highly effective way to build these structures where more traditional CNNs such as VGG struggle to achieve the same performance or make effective use of deep networks.  We have also shown that simulated channel effects, especially moderate \ac{LO} impairments improve the effect of transfer learning to \ac{OTA} signal evaluation performance, a topic which will require significant future investigation to optimize the synthetic impairment distributions used for training.  

\section{Conclusion}

\ac{DL} methods continue to show enormous promise in improving radio signal identification sensitivity and accuracy, especially for short-time observations.  We have shown deep networks to be increasingly effective when leveraging deep residual architectures and have shown that synthetically trained deep networks can be effectively transferred to over the air datasets with (in our case) a loss of around 7\% accuracy or directly trained effectively on \ac{OTA} data if enough training data is available.  While large well labeled datasets can often be difficult to obtain for such tasks today, and channel models can be difficult to match to real-world deployment conditions, we have quantified the real need to do so when training such systems and helped quantify the performance impact of doing so.

We still have much to learn about how to best curate datasets and training regimes for this class of systems.  However, we have demonstrated in this work that our approach provides roughly the same performance on high \ac{SNR} \ac{OTA} datasets as it does on the equivalent synthetic datasets, a major step towards real world use.  We have demonstrated that transfer learning can be effective, but have not yet been able to achieve equivalent performance to direct training on very large datasets by using transfer learning.  As simulation methods become better, and our ability to match synthetic datasets to real world data distributions improves, this gap will close and transfer learning will become and increasingly important tool when real data capture and labeling is difficult.  The performance trades shown in this work help shed light on these key parameters in data generation and training, hopefully helping increase understanding and focus future efforts on the optimization of such systems.

\printbibliography

\appendix 

\begin{figure*}[h]
    \centering
    \includegraphics[width=1.0\textwidth]{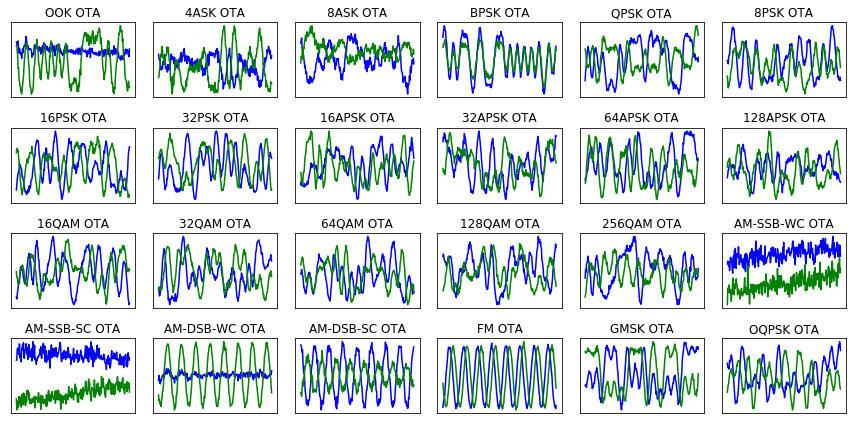}
    \caption{I/Q time domain examples of 24 modulations over the air at roughly 10 dB $E_s/N_0$ ($\ell = 256$) }
    \label{fig:const2}
\end{figure*}

\begin{figure*}[h]
    \centering
    \includegraphics[width=1.0\textwidth]{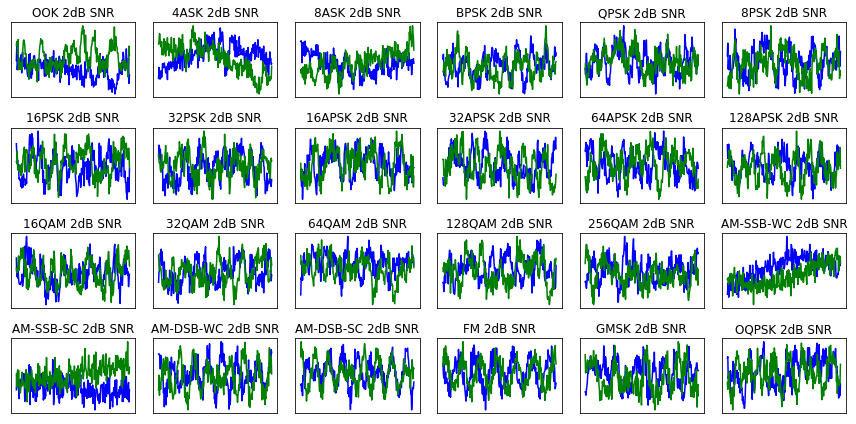}
    \caption{I/Q time domain examples of 24 modulations from synthetic $\sigma_{clk}=0.01$ dataset at 2dB $E_s/N_0$ ($\ell = 256$)}
    \label{fig:const4}
\end{figure*}

\end{document}